\documentclass[conference]{IEEEtran}
\IEEEoverridecommandlockouts
\usepackage{import}
\usepackage{url}

\usepackage{breakurl}
\usepackage[breaklinks]{hyperref}
\usepackage{tabularx}
\usepackage{multirow}
\usepackage{amsmath}
\usepackage{caption}
\usepackage{subfig}
\usepackage{tikz}
  \usetikzlibrary{positioning}
  \usetikzlibrary{tikzmark} % arrows in tex
  \usetikzlibrary{arrows}   % arrows in tex
  \usetikzlibrary{calc}     % (node)+(3cm,2cm)
  \tikzstyle{every picture}+=[remember picture]

\makeatletter
\def\endthebibliography{%
	\def\@noitemerr{\@latex@warning{Empty `thebibliography' environment}}%
	\endlist
}
\makeatother

\usepackage{cite}
\usepackage{amsmath,amssymb,amsfonts}
\usepackage{algorithmic}
\usepackage{booktabs}
\usepackage{graphicx}
\usepackage{subcaption}
\usepackage{textcomp}
\usepackage{xcolor}
\usepackage[left=0.75in,right=0.75in,top=1in,bottom=0.75in]{geometry}
\def\BibTeX{{\rm B\kern-.05em{\sc i\kern-.025em b}\kern-.08em
    T\kern-.1667em\lower.7ex\hbox{E}\kern-.125emX}}
\begin{document}

\title{\LARGE \bf CAPS: Context-Aware Priority Sampling for Enhanced \\ Imitation Learning in Autonomous Driving
}

\author{
Hamidreza Mirkhani\textsuperscript{\rm 1}, Behzad Khamidehi\textsuperscript{\rm 1}, Ehsan Ahmadi\textsuperscript{\rm 1,2}, Mohammed Elmahgiubi\textsuperscript{\rm 1}, \\Weize Zhang\textsuperscript{\rm 1}, Fazel Arasteh\textsuperscript{\rm 1}, Umar Rajguru\textsuperscript{\rm 1}, Kasra Rezaee\textsuperscript{\rm 1}, Dongfeng Bai\textsuperscript{\rm 1}\\
{\textit{\textsuperscript{\rm 1}Noah's Ark Lab, Huawei Technologies Canada} \quad  \textit{\textsuperscript{\rm 2}University of Alberta}} \\
{\textit{Emails: firstname.lastname@huawei.com}} \\%
% \normalsize $^{*}$These authors contributed equally \\% <-this % stops a space
}

\maketitle

\begin{abstract}
In this paper, we introduce Context-Aware Priority Sampling (CAPS), a novel method designed to enhance data efficiency in learning-based autonomous driving systems. CAPS addresses the challenge of imbalanced datasets in imitation learning by leveraging Vector Quantized Variational Autoencoders (VQ-VAEs). In this way, we can get structured and interpretable data representations, which help to reveal meaningful patterns in the data. These patterns are used to group the data into clusters, with each sample being assigned a cluster ID. The cluster IDs are then used to re-balance the dataset, ensuring that rare yet valuable samples receive higher priority during training. We evaluate our method through closed-loop experiments in the CARLA simulator. The results on Bench2Drive scenarios demonstrate the effectiveness of CAPS in enhancing model generalization, with substantial improvements in both driving score and success rate.

\end{abstract}


\section{Introduction}
\label{sec:introduction}

Imitation learning (IL) is a widely used approach for end-to-end training in autonomous driving (AD), 
where policies are learned from expert demonstrations.
However, a significant challenge arises from the nature of expert datasets: they predominantly consist of trivial scenarios, including cruising in a straight line and decelerating at a stop sign, which can be easily handled even by a rule-based planner. Meanwhile, edge cases, such as parking cut-ins, sudden stops, and near-accident incidents, are rare and challenging to handle, and they might be overshadowed by the trivial scenarios as their frequencies are much lower.

\begin{figure}
    \centering
    \includegraphics[clip, trim=0.0cm 0.0cm 0.0cm 0.0cm, width=1.0\linewidth]{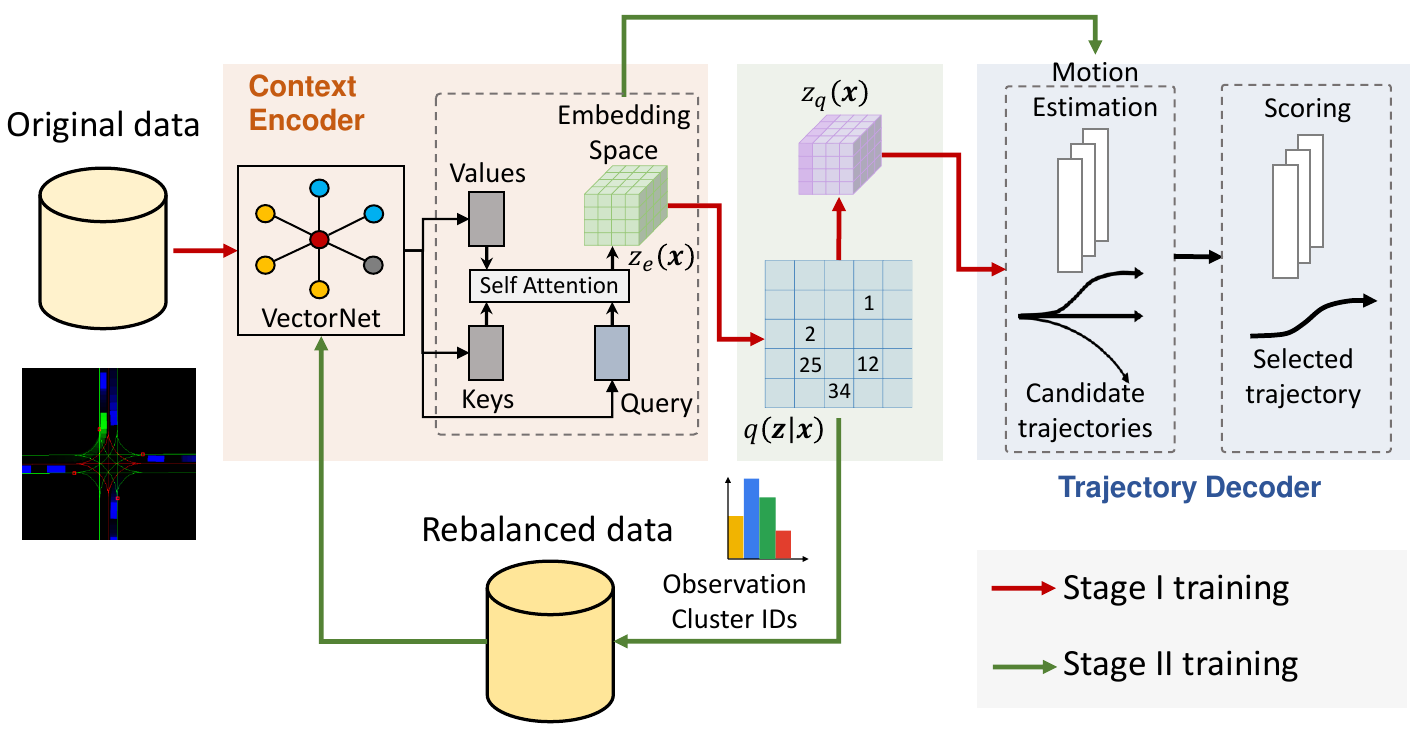} \\
    \vspace*{-0.0cm}
    \caption{Block diagram of the CAPS framework. In Stage I, a VQ-VAE module is introduced to facilitate the joint training of the clustering model and the planner. In Stage II, the trained VQ-VAE generates codebook IDs for the training samples, which are re-weighted based on cluster frequencies to refine the main planner.}
    \vspace*{-0.0cm}
    \label{fig:overview}
\end{figure}

The data imbalance leads to IL models overfitting to common driving behaviors while failing to generalize to rare but critical edge cases \cite{distributional-shift-ross2011reduction, distributional-shift-pmlr-v9-ross10a}. Naively scaling up the dataset size, even though it is costly and time-consuming, is not efficient, as the trivial scenarios with minimal learning value are still dominating the dataset. As highlighted in \cite{bronstein2022embedding}, an agent trained on just $10\%$ of the data can perform as well as one trained on the full dataset, 
underscoring that a substantial portion of uniformly collected data does not meaningfully contribute to learning robust driving policies. This data imbalance becomes even more problematic in closed-loop evaluation: 
while a planner trained on a uniformly sampled dataset may perform well in expectation, 
a single failure in an edge case can lead to catastrophic consequences \cite{NEURIPS2019_94701864, 10611038}. Thus, ensuring strong generalization to rare, high-risk situations is more critical than merely excelling in routine driving scenarios.

Data balancing is a potential solution to the data imbalance problem. However, this requires access to labels for the samples, which requires costly human labeling, which is not well-scalable to large datasets. Furthermore, the criteria given to the annotators to categorize the samples are subjective and depend on the application domain and the task at hand, and need to be changed with the changes of the objective. For example, the class labels that are used for the data balancing in the trajectory prediction task is different from the class labels that are used for the data balancing in the planning task, as in the trajectory prediction task the focus in on minimizing the prediction error in trajectory space, while in the planning task the focus is on maximizing the safety of the driving behavior even if it deviates from logged expert's demonstrations, e.g. having less human-like driving behavior.

Another possible solution is to use rule-based approaches, which do not require human labour for data annotation.
For instance, we can apply k-nearest neighbors (KNN) clustering in the trajectory space to cluster the samples into different classes, and then use the cluster frequencies to re-weight the samples. However, this approach skips the context information, and it lacks the capacity to discriminate complex nuances in the dataset. 
For example, trajectory-based clustering cannot differentiate between a scenario where the ego vehicle is decelerating due to reaching a red light and a scenario where the ego vehicle is decelerating as a result of observing a rare collision ahead. Similarly, in trajectory prediction, models that neglect causal inter-agent context have been shown to suffer from poor robustness and generalization \cite{Ahmadi_causal}.

To have a more effective scenario labeling to be used for data balancing, we propose CAPS (Context-Aware Priority Sampling), 
a novel framework that leverages variational self-supervised representation learning. 
CAPS adopts Vector Quantized Variational Autoencoders (VQ-VAE) \cite{NIPS2017_7a98af17} to conditionally encode the future trajectory of the ego-vehicle into a quantized latent representation given the scene context information. 
As a byproduct of the VQ-VAE trajectory encoding, the expert demonstrations are automatically clustered into several distinct classes.
This approach is different from the traditional clustering approaches that focus solely on trajectory features, as CAPS captures rich contextual information that is effective in reconstructing the future trajectory of the ego-vehicle.

CAPS, similar to the recent generative models \cite{esser2021taming,ramesh2021zero}, employs a two-stage training process, 
In stage I, first, the model is trained to encode the contextful future trajectory of the ego-vehicle into a quantized latent representation, and use the quantized codebook indices as the clustering labels for the data samples.
We assign a sampling weight to each data sample based on the inverse of its cluster frequency.
In stage II, which is the main training stage, we use a weighted sampling strategy to train the planner model with an enhanced focus on the underrepresented data samples.
The training of the stages is decoupled, which allows us to focus on the quality of the representation learning and have a good clustering model regardless of the objective functions of the downstream task, here, the planning task. In stage II training, we only need the class-balanced weights from stage I, and the focus is on the performance of the planner model without missing the underrepresented scenario classes.

To summarize, we can highlight the following contributions of this work:
\begin{itemize}
    \item We introduce CAPS (Context-Aware Priority Sampling), a novel framework designed to effectively learn a context-aware representation of the expert demonstrations, which is used for a class-balanced training for the planning task. 
    \item We conduct extensive closed-loop experiments on the Bench2Drive benchmark dataset \cite{benchdrive-jia2024}, demonstrating that our proposed CAPS framework consistently outperforms baseline methods. In particular, CAPS surpasses trajectory endpoint and anchor-based clustering approaches, which represent intuitive rule-based strategies for trajectory clustering. Furthermore, we show that CAPS achieves superior performance compared to other state-of-the-art methods with a similar computational budget in the planning task.
\end{itemize}

In Section \ref{sec:method}, we discuss the implementation details of CAPS.
In Section \ref{sec:results}, we demonstrate that CAPS achieves roughly a 10\% improvement in Driving Score and Success Rate on Bench2Drive scenarios, illustrating the benefits of context-aware priority sampling for autonomous driving without the need for additional expert data or extra computational cost during deployment.
\section{Related Works}
\label{sec:related_works}

Prior approaches on data balancing focus on modifying demonstration data before training through various pre-processing techniques, such as up-sampling minority clusters or down-sampling majority clusters. Many of these methods rely on SMOTE \cite{SMOTE}, which generates synthetic samples for the minority class by interpolating between minority samples and their $k$ nearest neighbors in the latent space. For instance, studies in \cite{TEMRAZ2022100375, KOZIARSKI2020106223, Lei2020} explore up-sampling by creating synthetic instances for minority classes, while \cite{NIU2020120, GUZMANPONCE2021114301} focus on down-sampling the majority class to balance the dataset. Another common strategy is data augmentation, which expands the training set through synthetic transformations \cite{10588830, chauffeurnet, ning2023input}. For example, \cite{chauffeurnet} perturbs demonstration trajectories by altering their midpoints with a uniform distribution, while keeping the start and end points intact. Although these techniques are sample-efficient and require fewer computational resources, they often depend on offline clustering or manual labeling to determine which samples should be perturbed or augmented—a process that becomes increasingly impractical and time-consuming as dataset size grows.

To mitigate dataset limitations during deployment, another group of methods leverages expert policies to improve closed-loop performance. The key idea is to continuously collect additional samples from both the expert and the trained policy, in contrast to behavior cloning \cite{9117169}, which relies on a fixed set of demonstrations. These methods dynamically update the dataset with states encountered by the learner during execution \cite{distributional-shift-ross2011reduction}, but require collecting large numbers of expert-labeled samples, making them resource-intensive. Several works attempt to address these challenges \cite{zhang2016queryefficientimitationlearningendtoend, Bicer_2019, menda2019ensembledaggerbayesianapproachsafe}. For example, SafeDAgger \cite{zhang2016queryefficientimitationlearningendtoend} predicts unsafe trajectories and queries the expert only in those cases. While such approaches offer improvements in closed-loop performance, they still rely heavily on expert supervision, which limits their practicality. Moreover, they often depend on planner failures as a trigger for identifying valuable training data, which is inefficient and can result in suboptimal learning.

More recent work combines imitation learning with reinforcement learning to further enhance robustness. In these approaches, RL rewards guide models initially trained on demonstration data \cite{lu2023imitation, bronstein2022hierarchical, booher2024cimrl, liu2023blending}. For example, \cite{lu2023imitation} presents a hybrid method that leverages large-scale human driving data through IL and incorporates RL to handle rare and challenging scenarios. Similarly, \cite{bronstein2022hierarchical} introduces a hierarchical model that blends IL and RL, improving generalization to unseen situations and navigation in complex environments. Despite their promise, these approaches typically require vast amounts of data and computational resources, which can be impractical in real-world scenarios, particularly given the difficulty of achieving realistic autonomous driving simulations \cite{SMART_2, ahmadi2024rlftsim}.

Taken together, these methods highlight the need for solutions that can automatically identify and prioritize the most informative data samples for training and fine-tuning. Instead of relying on failures to signal valuable data, an effective framework should analyze the encoded context to proactively recognize infrequent yet critical cases. Prioritizing such samples during training can reduce manual effort, improve sample efficiency, and lead to more robust learning outcomes.

\section{Methodology}
\label{sec:method}
A key challenge in training motion planning models is to determine which data samples most effectively contribute to improving closed-loop performance. 
One approach to address this is clustering data based on the main characteristics of the ego trajectory, such as position, speed, acceleration, curvature, etc. However, if this clustering is based solely on the ego vehicle's trajectory features, it may overlook crucial contextual information from the surrounding environment. Closed-loop performance is influenced not only by the ego trajectory but also by interactions with other agents and environmental factors. Ignoring this context can result in suboptimal outcomes during closed-loop tests. A holistic evaluation should therefore assess each sample by considering both the key features of the ego trajectory and the surrounding context.
Unlike traditional VAEs that rely on continuous latent variables, VQ-VAE learns a finite set of latent codes, enabling a more structured and interpretable data representation \cite{NIPS2017_7a98af17}. By encoding contextual information, VQ-VAE ensures that similar inputs are mapped to nearby or identical latent codes, effectively capturing meaningful patterns and dependencies. This discrete representation is particularly beneficial as it reduces sensitivity to small variations, making clustering more robust and ensuring that learned representations remain stable across different inputs. Additionally, discrete latent spaces help mitigate posterior collapse, a common issue in VAEs where the latent space becomes less informative.

We employ a two-stage training process. In Stage 1, we train both VQ-VAE and our planner, where the planner operates in a generative mode to produce trajectories as outputs. Importantly, our planner is trained with the clustering model integrated into the training loop, ensuring that trajectory generation aligns with the learned discrete representations. Once both models are trained, Stage 2 deploys the clustering model to assign a cluster ID to each training sample. We then apply sample weighting based on cluster frequencies to balance the distribution of the training data for the planner, enhancing its ability to generalize across different scenarios. 

\subsection{Model Architecture}\label{model-architecture}
Figure \ref{fig:overview} illustrates the architecture of our model, which integrates both the planner and the clustering modules. We describe each component in more detail as follows. 

\noindent\textbf{Context Encoder:}
The planner is composed of a context encoder and a trajectory decoder. For the context encoder, we employ VectorNet \cite{9157331}, which processes the observed and future states of the ego vehicle, surrounding objects, and map information. To simplify the notation, we use subscripts \( p \) and \( f \) to denote \textit{past} and \textit{future}, respectively. Specifically, \( \mathbf{s}^{\text{ego}}_{\text{p}} \) and \( \mathbf{s}^{\text{ego}}_{\text{f}} \) represent the past and future state sequences of the ego vehicle, while \( \mathbf{s}^{\text{obj}}_{\text{p}} \) and \( \mathbf{s}^{\text{obj}}_{\text{f}} \) correspond to the past and future states of surrounding objects. Additionally, \( \mathbf{c}_{\text{p}} \) and \(\mathbf{c}_{\text{f}} \) capture the past and future context of the scene. With these definitions, the input to the scene encoder is represented as:
\begin{equation}
    \mathbf{x} = \left\{ (\mathbf{s}^{\text{ego}}_{\text{p}}, \mathbf{s}^{\text{ego}}_{\text{f}}), (\mathbf{s}^{\text{obj}}_{\text{p}}, \mathbf{s}^{\text{obj}}_{\text{f}}), (\mathbf{c}_{\text{p}}, \mathbf{c}_{\text{f}}) \right\}.
\end{equation}\label{stage-1-inputs-1}

The embedded features are then passed to a transformer module to integrate information from all components of the scene. This allows the system to model the relationships between agents, road elements, and surrounding objects. We employ a multi-head attention module to enable the model to focus on multiple aspects of the scene simultaneously, learning diverse patterns of interaction that are essential for accurate decision-making in dynamic environments. 

\noindent\textbf{Trajectory Decoder:} Our trajectory decoder module is designed based on \cite{Zhao_TNT}, adapting its structure to better suit our planning framework. It consists of two key components: a trajectory generation module and a scoring module. The trajectory generation module produces a diverse set of candidate future trajectories, each conditioned on selected target points. These targets represent plausible endpoints within a predefined time horizon, ensuring that the generated trajectories align with feasible motion patterns in real-world driving scenarios. 
The scoring module evaluates and ranks these trajectories based on their plausibility, selecting the most suitable one for execution. To adapt this architecture for planning applications, we incorporate contingency masks, as introduced by \cite{arasteh2024validitylearningfailuresmitigating}. These masks help mitigate compounding errors during sequential decision-making by filtering out unrealistic or unsafe trajectory candidates, leading to more reliable and stable planning in dynamic environments.

\noindent\textbf{Clustering Module:} 
Our clustering module is based on a VQ-VAE architecture, as shown in Fig. \ref{fig:overview}, enabling the discrete representation of continuous data. VQ-VAE quantizes the embedding space using a codebook, assigning each data sample to a cluster. Let $\mathbf{z}_{e}(\mathbf{x})$ denote the embedding representation of input $\mathbf{x}$, obtained from the context encoder module.
The embedding space $\mathbf{z}_{e}(\mathbf{x})$ has dimensions $N_{\text{obj}} \times d_{e}$, where $N_{\text{obj}}$ is the total number of objects in the scene, and $d_e$ is the dimension of latent vectors. The first column of the embedding space corresponds to the ego vehicle's feature, encapsulating critical scene context from the ego's perspective. We denote this vector as $\mathbf{z}^{\text{ego}}_{e}$ and use it as input to our VQ module. The codebook consists of $K$ embedding vectors $\{e_1, e_2, \dots, e_K\}$, where each $e_j \in \mathbb{R}^{d_e}$ represents a discrete latent vector. Following the VQ-VAE pipeline \cite{NIPS2017_7a98af17}, we quantize $\mathbf{z}^{\text{ego}}_{e}$ by finding the closest discrete latent variable $e_k$ from the codebook:
\begin{equation} 
k = \arg\min_{j} || \mathbf{z}^{\text{ego}}_{e} - e_j ||_2. 
\end{equation}
The quantized representation, which serves as input to the VQ decoder, is then given by $\mathbf{z}^{\text{ego}}_{e} = e_k$. The decoder output is used to reconstruct the input, ensuring meaningful latent representations. The losses used for training are introduced in the next section.

\subsection{Model Training Approach} \label{training-stages}

As previously described, our pipeline follows a two-stage training process. In the first stage, we jointly train the planner and the VQ-VAE model to determine the clustering labels for each data sample. During this phase, the planner functions as a generative model, ensuring that the model's input and output remain consistent while learning meaningful embedding representations. For the planner, we employ an imitation loss similar to \cite{Zhao_TNT}, which aims to reconstruct the ego trajectory in the output. For the VQ model, we optimize the following loss function

\begin{equation}
    \begin{aligned}
        \mathcal{L} = &  \log p(\mathbf{z}_e^{\text{ego}}|\mathbf{z}_q^{\text{ego}}) + \\
        & | sg([\mathbf{z}_e^{\text{ego}}] - e |^2 + \beta | \mathbf{z}_e^{\text{ego}} - sg[e] |^2.
    \end{aligned}
\end{equation}
The first term represents the reconstruction loss, which ensures that the encoder and decoder accurately reconstruct the ego features. The second term, known as the codebook loss, minimizes the discrepancy between the encoder’s output and the closest codebook vector. Here, $sg[.]$ denotes the stop-gradient operation, which prevents direct updates to the codebook while encouraging the encoder to align with the learned embeddings. Since the embedding space has no inherent scale, its magnitude can grow indefinitely if the embeddings $e$ fail to update at the same rate as the encoder parameters. To address this, the third term, referred to as the commitment loss (introduced by \cite{NIPS2017_7a98af17}), ensures that the encoder consistently maps inputs to a specific embedding, preventing uncontrolled expansion. 

In the second stage, we leverage the trained VQ-VAE model to assign a cluster label to each training sample. Based on cluster frequencies, we then adjust sample weights to balance the training data for the planner. The planner is subsequently trained using an imitation loss similar to the one proposed in \cite{Zhao_TNT}, which aims to reconstruct the ego trajectory in the output. This approach improves the planner’s ability to generalize across diverse scenarios by ensuring a more balanced representation of training samples across different clusters.
\section{Results and Discussions}
\label{sec:results}

\subsection{Experiments Setup}

\begin{figure*}
    % \captionsetup{font=scriptsize}
    \vspace*{-0.0cm}
    \centering
    \includegraphics[clip, trim=0.0cm 13.2cm 4.0cm 3.8cm, width=1.0\linewidth]{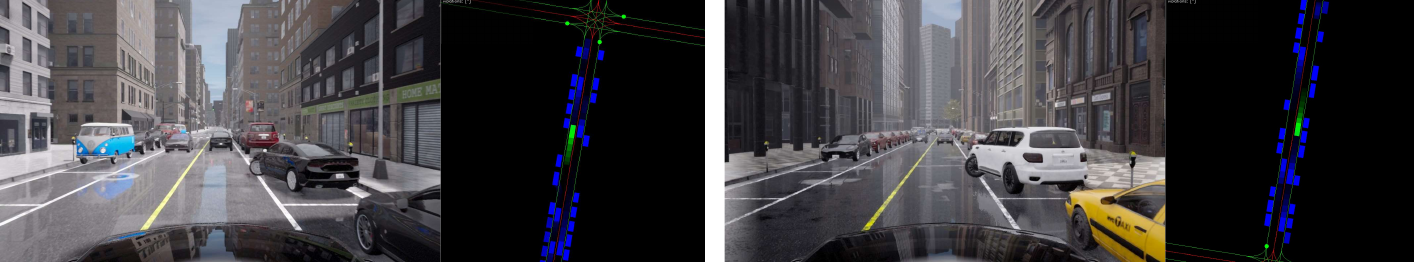}
    \includegraphics[clip, trim=0.0cm 13.3cm 4.0cm 3.8cm, width=1.0\linewidth]{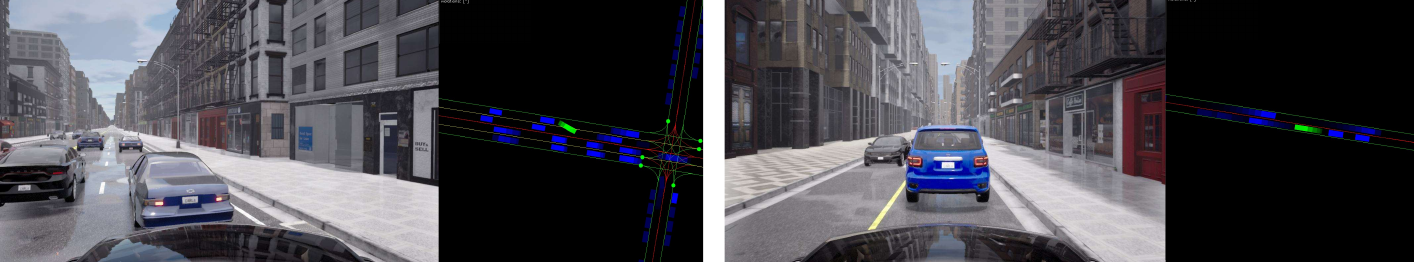}
    % \includegraphics[clip, trim=0.2cm 9.8cm 8.0cm 6.0cm, width=1.0\linewidth]{images/results-01-model-B.pdf}\\
    % \vspace*{-0.5cm}
    \caption{Comparison of two examples of scenes clustered using our approach, where each row corresponds to the same codebook ID. The first row predominantly contains parking cut-in scenarios, while the second row features instances where the ego vehicle stops and waits behind a stationary vehicle or obstacle. In the bird's-eye-view images of the traffic scenes, the ego vehicle and the surrounding vehicles are shown in green and blue, respectively.}
    % \vspace{-0.2cm}
    \label{fig:results-snapshots-1}
\end{figure*}

\noindent \textbf{Simulation Environment:} We use CARLA Leaderboard 2.0 for all our experiments. CARLA Simulator \cite{carla-open-urban-driving-2017} provides a highly detailed virtual environment to design and train AI models in tasks such as perception, decision-making, and control. CARLA Leaderboard 2.0 extends the original leaderboard by providing a standardized set of tasks and evaluation metrics for autonomous driving systems. It encompasses a diverse range of scenarios and challenges designed to assess multiple aspects of driving performance.\par
\noindent\textbf{Data Collection:} One major challenge in using CARLA Leaderboard 2.0 is that no single expert can consistently perform across its diverse scenarios. Inspired by \cite{benchdrive-jia2024}, we developed an automated scenario generation module that segments each route in Leaderboard 2.0 into multiple short-duration segments, with each segment representing a distinct scenario. This approach allows us to compile a more balanced dataset for training our planner and efficiently filter out failing scenarios, compared to collecting data from full routes. Additionally, to further enhance our training data, we incorporate high-quality data provided by the official Leaderboard 2.0 team. Their CARLA logs feature manual execution of each scenario, all achieving a $100\%$ score, and include pre-calculated vehicle trajectories for all elements within the scenario.\par
\noindent\textbf{Evaluation Benchmark:} We use Bench2Drive \cite{benchdrive-jia2024} for benchmarking. It provides a comprehensive set of 220 short-segment scenarios that capture various challenging traffic behaviors. \par
\noindent\textbf{Evaluation Metrics:} We focus on two key metrics reported in Bench2Drive:
\begin{itemize}
    \item \textit{Driving Score}: The driving score evaluates the quality of driving behavior by penalizing infractions such as collisions, off-road driving, and failure to adhere to traffic rules, as defined in the official CARLA Leaderboard 2.0.
    \item \textit{Success Rate}: The success rate measures the percentage of scenarios in which the vehicle successfully reaches its goal without experiencing critical failures. A critical failure is defined as any infraction (with the exception of 'min-speed infraction') captured by the CARLA Simulator. 
\end{itemize}
\par

\subsection{Experimental Results}
We evaluate our approach on CARLA using both sensor inputs and privileged inputs.
Privileged inputs are defined as a vectorized scene representation that is accessible by CARLA. 
In the experiment with the privileged inputs, 
the effect of the perception module is isolated, and the planner has direct access to a more compact and accurate representation of the environment.
For sensor-based experiments, we deploy six cameras around the vehicle and fine-tune a version of the VAD backbone to estimate the vectorized scene representation \cite{vad-jiang2023, huawei-cvpr-carla-2024}. Our training dataset consists of approximately $385k$ samples collected using our rule-based planner agent, combined with around $178k$ samples obtained by replaying logs from the official CARLA Leaderboard 2.0 team. In our experiment, we set the number of codebook IDs to 64. As previously discussed, in the VQ-VAE implementation, the codebook IDs are assigned to the quantized embedding vectors in the latent space, which also serves as the cluster ID for data samples used in Stage I of our pipeline.

\begin{table}[t]
\raggedright
\caption{Planning performance for data balancing with different clustering methods. The results are presented for 220 short segment scenarios of Bench2Drive.}
\label{table_clustering}
\begin{tabular}{
    >{\raggedright\arraybackslash}m{1cm}
    >{\raggedright\arraybackslash}m{4.0cm}
    >{\raggedright\arraybackslash}m{0.9cm}
    >{\raggedright\arraybackslash}m{0.9cm}
}
\noalign{\hrule height 0.05cm}
\textbf{Input} & \textbf{Clustering Method} & \textbf{Driving Score$\uparrow$} & \textbf{Success Rate(\%)$\uparrow$} \\
\hline
\multirow{3}{1cm}{Privileged} 
& End-point & 59.63 & 48.23 \\
& Anchor-based & 62.60 & 51.83 \\
& PER & 61.18 & 50.26 \\
& CAPS & \textbf{68.91} & \textbf{56.97} \\
\hline
\multirow{3}{8cm}{Sensor} 
& End-point & 57.06 & 46.16 \\ 
& Anchor-based & 60.64 & 48.69 \\ 
& PER & 60.42 & 47.98 \\
& CAPS & \textbf{66.76} & \textbf{52.87} \\ 
\noalign{\hrule height 0.05cm}
\end{tabular}
\label{tab:grouped_results}
\end{table}

Table \ref{table_clustering} reports the closed-loop performance of our planner across different clustering and rebalancing strategies on $220$ short-segment scenarios from Bench2Drive \cite{benchdrive-jia2024}, considering both privileged and sensor inputs. We evaluate two rule-based clustering strategies: endpoint clustering, where the ego trajectory is represented by its start and end points, and anchor-based clustering, which uses the full 3-second ego trajectory. Both methods employ k-means to partition the dataset into 64 clusters, providing a consistent comparison baseline. In contrast, CAPS encodes both the ego trajectory and its surrounding context in the scene embeddings while maintaining the same number of clusters. Results show that CAPS consistently outperforms the rule-based approaches, emphasizing the value of context-aware representations for planner performance.
We also investigate Prioritized Experience Replay (PER) sampling \cite{per}, where training samples are weighted by their loss values, and higher-loss samples are sampled more frequently. We find that this strategy does not significantly improve imitation learning performance, as planner training should prioritize closed-loop metrics such as success rate and driving score. Adjusting sample weights based on open-loop metrics alone provides no measurable benefit, as confirmed by our evaluation results.

\begin{table}[t]
\centering
\caption{Effect of contextual information during clustering. Removing the surrounding context increases the time required to complete scenarios despite similar infraction statistics.}
\label{table_context_ablation}
\resizebox{\linewidth}{!}{%
\begin{tabular}{lcccc}
\noalign{\hrule height 0.05cm}
\textbf{Model} &
\begin{tabular}[c]{@{}l@{}}\textbf{Avg. Comp. }\\\textbf{Time (s)}\textbf{$\downarrow$}\end{tabular} &
\begin{tabular}[c]{@{}l@{}}\textbf{Route }\\\textbf{Comp.}\textbf{(\%)$\uparrow$}\end{tabular} &
\begin{tabular}[c]{@{}l@{}}\textbf{Success }\\\textbf{Rate}\textbf{(\%)$\uparrow$}\end{tabular} &
\begin{tabular}[c]{@{}l@{}}\textbf{Driving}\\\textbf{Score}\textbf{$\uparrow$}\end{tabular} \\
\midrule
No Clustering & 71.4 & 92.3 & 48.2 & 59.5 \\
No-Agent Context & 69.3 & 90.1 & 51.8 & 66.6 \\
No-Agent/Map Context & 74.4 & 89.8 & 49.7 & 63.8 \\
CAPS & \textbf{50.0} & \textbf{97.3} & \textbf{52.9} & \textbf{66.7} \\\noalign{\hrule height 0.05cm}
\end{tabular}
}
\end{table}

To further analyze the importance of contextual information in the clustering stage, we perform additional context-exclusion ablations during Stage I training. The results of these studies are reported in Table~\ref{table_context_ablation}, including: i) \textbf{No-Agent Context}, where we excluded the all of the agents from the scene, so clustering is done using ego trajectory and the map elements, ii) \textbf{No-Agent/Map Context}, where we excluded both agents and the map elements from the scene. Regarding the evaluation metrics, we report average scenario completion time, route completion percentage, success rate, and driving score. According to the results, CAPS outperforms the ablated models and the baseline in all metrics. While the gain in the driving score metric is relatively small, we observe that there is a larger improvement in the other metrics; the scenario completion time for CAPS is 32\% lower than No-Agent/Map, and it is 28\% lower than the No-Agent model. Similarly, for the route completion metric, CAPS gets major improvements.
These results indicate that incorporating scene context during clustering leads to improved performance of the learned planner.

\begin{figure*}[t]  % * for spanning both columns
    \centering
    % First row: single image (full width)
    \subfloat[]{%
        \includegraphics[clip, trim=0.0cm 0cm 0cm 0cm, width=1.0\linewidth]{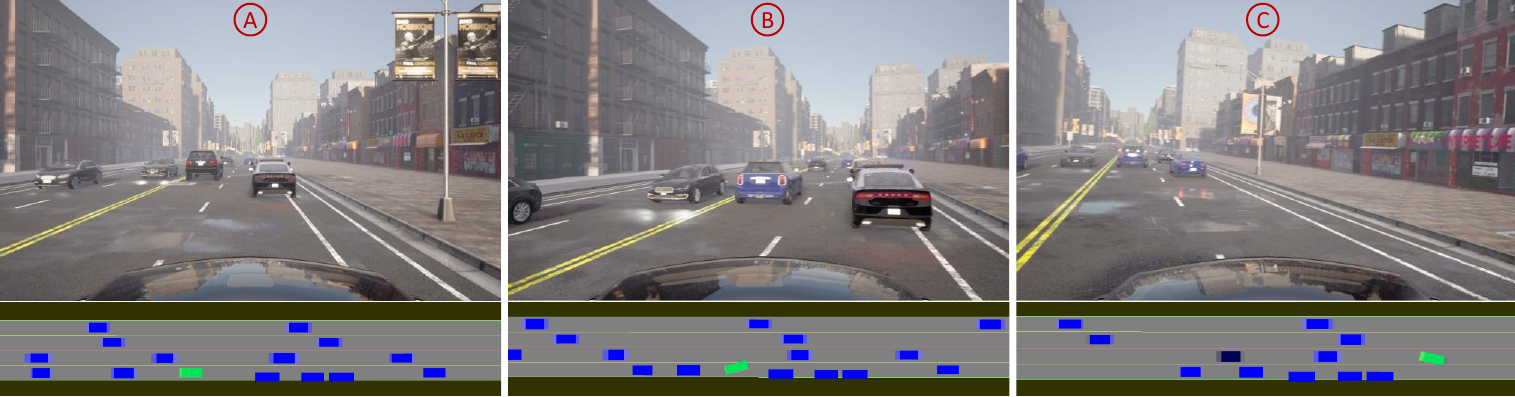}%
        \label{fig:sub1}}%

    % Second row: two images, each column width
    \subfloat[]{%
        \includegraphics[width=\columnwidth,trim=0 0 0 0,clip]{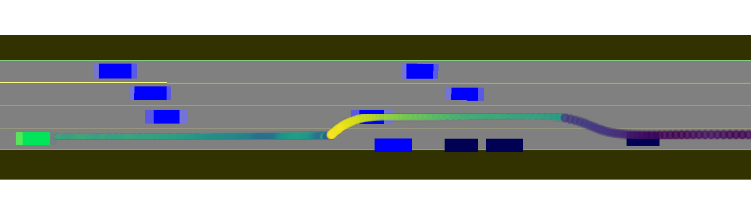}%
        \label{fig:sub2}}%
    \hfill
    \subfloat[]{%
        \includegraphics[width=\columnwidth,trim=0 0 0 0,clip]{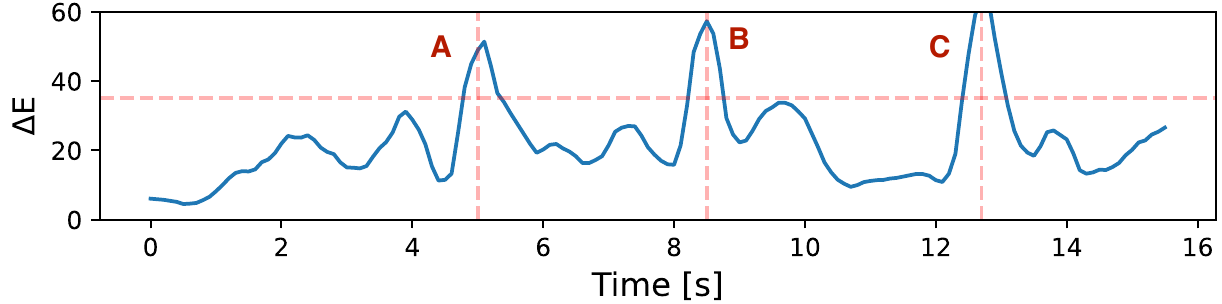}%
        \label{fig:sub3}}%
    
    \caption{ Qualitative temporal probing of a driving scenario with CAPS. (a) Key transitions in a driving scenario are captured by the changes in the CAPS's quantized embedding space. The ego vehicle encounters lane congestion due to an accident and undergoes three main transitions: (A) slowing down and negotiating with adjacent vehicles, (B) changing lanes to bypass the blockage, and (C) returning to the right lane after passing the accident. (b) The ego trajectory is colored according to its cluster ID. To obtain these colors, the scene embeddings are first reduced using t-SNE, and then each point is assigned a color based on its cluster.  (c) Temporal embedding distance of the scene. Sudden jumps in embedding distance (vertical dashed lines in this figure) correspond to these key frames.}
    \label{fig:traj-with-tsne}
\end{figure*}

To provide additional evidence of the effectiveness of our approach in clustering scenes, we present different samples with the same cluster IDs in Fig. \ref{fig:results-snapshots-1}. Each row in this figure corresponds to a distinct cluster ID, highlighting the consistency of the cluster characteristics across different scenes. The first row presents scene samples corresponding to the parking cut-in scenario, while the second row shows scenarios where the ego vehicle stops and waits behind a stationary vehicle or obstacle. Despite coming from different scenes, the nature of the scenarios within the same cluster is similar.

To examine temporal consistency and continuity across consecutive frames, we select a segment of a driving scenario where the ego vehicle travels along a two-way, double-lane road until it encounters lane congestion caused by an accident (Fig. \ref{fig:traj-with-tsne}). Because the scene contains multiple dynamic agents, we highlight only the most distinguishable key frames—those where a human expert can intuitively map the evolving situation to the transitions that our method identifies as critical. By probing the temporal changes of the embedding space and retrieving the camera frames associated with the large changes, we observe three key moments: 

\begin{enumerate}
    \item when the ego vehicle first perceives the congestion, decelerates sharply to avoid collision, and begins negotiating with vehicles in the adjacent lane to bypass the blockage (Fig. \ref{fig:traj-with-tsne})-A);
    \item when a sufficient gap opens in the left lane, allowing the ego vehicle to change lanes safely (Fig. \ref{fig:traj-with-tsne})-B);
    \item when the ego vehicle clears the accident scene and attempts to merge back into the right lane (Fig. \ref{fig:traj-with-tsne})-C).
\end{enumerate}

Ideally, the codebook IDs and their corresponding embeddings before and after these transitions should capture the changes in the scene. Fig. \ref{fig:traj-with-tsne}) illustrates this by showing the temporal differences in the embedding space produced by the VQ-VAE module. Each sudden jump in embedding distance reflects a key transition. We also note sporadic jumps in the temporal embedding distances, resulting from CARLA abruptly spawning vehicles, which induce significant scene and cluster ID changes.

\begin{table}[t]
\raggedright
\caption{ Planning performance benchmarking with closed-loop simulation in CARLA. The results are presented for
220 short segment scenarios of Bench2Drive. Comparison is done with other approaches using similar computational budgets.}
\label{table_trained_constraint}
\begin{tabular}{
    >{\raggedright\arraybackslash}m{0.8cm}
    >{\raggedright\arraybackslash}m{4.5cm}
    >{\raggedright\arraybackslash}m{0.9cm}
    >{\raggedright\arraybackslash}m{0.9cm}
}
\noalign{\hrule height 0.05cm}
\textbf{Input} & \textbf{Method} & \textbf{Driving Score$\uparrow$} & \textbf{Success Rate(\%)$\uparrow$} \\
\hline
\multirow{3}{1cm}{Privileged} 
& IL with Uniform Sampling & 62.26 & 54.16 \\
& CAPS (Ours) & 68.91 & 56.97 \\ 
& Expert & 82.37 & 84.29  \\
\hline
\multirow{10}{8cm}{Sensor} 
& AD-MLP\cite{zhai2023rethinking} & 18.05 & 00.00 \\ 
& UniAD-Base\cite{hu2023planning} & 45.81 & 16.36 \\ 
& VAD\cite{vad-jiang2023} & 42.35 & 15.00 \\ 
& TCP-traj\cite{wu2022trajectory} & 59.90 & 30.00 \\ 
& ThinkTwice\cite{jia2023think} & 62.44 & 31.23 \\ 
& DriveAdapter\cite{jia2023driveadapter} & 64.22 & 33.08 \\ 
& IL with Uniform Sampling & 59.49 & 48.16 \\ 
& CAPS (Ours) & \textbf{66.76} & \textbf{52.87} \\ 
& Expert & 75.82 & 82.72  \\
\noalign{\hrule height 0.05cm}
\end{tabular}
\label{tab:grouped_results}
\end{table}
Table \ref{table_trained_constraint} compares the closed-loop performance of our planner trained using IL with other advanced end-to-end planners. We also include a comparison with an expert classical planner—a rule-based planner manually fine-tuned using privileged perception data from CARLA \cite{weize_peyman_iv, huawei-cvpr-carla-2024}. While the planner trained without CAPS demonstrates reasonable performance, its performance does not surpass the baselines. By incorporating CAPS into the training pipeline and rebalancing the dataset, however, the same planner consistently outperforms all baselines. It highlights the effectiveness of our approach in prioritizing the most valuable samples. It is worth noting that some recent approaches, such as SimLingo (CarLLaVa) \cite{simlingo} could achieve higher performance than our base planner models. However, these models operate in a different scope and rely on much larger vision-language models (VLMs), with up to 100$\times$ more parameters, which require significantly higher training/inference resources. Therefore, we excluded VLM-based approaches to ensure a fair comparison. 

\section{Conclusions}
\label{sec:conclusions}
In this study, we introduced a context-aware priority sampling framework and demonstrated its effectiveness in enhancing the performance of a learning-based motion planner. By prioritizing rare and challenging samples that may otherwise be underrepresented during training, the framework improves sample efficiency and closed-loop robustness. Beyond planner training, our approach can also be applied during the data-collection stage to selectively capture high-value driving experiences, thereby reducing the storage and processing of redundant or low-information data. This is particularly important given that a single autonomous vehicle can generate terabytes of raw data annually, making real-time identification of the most informative samples critical for scalable fleet-wide training \cite{activead-lu2024, greer2024whywhenuseactive}. Future research directions include investigating alternative VQ-VAE architectures such as \cite{fsq-mentzer2024finite}, integrating the framework with closed-loop training pipelines as in \cite{arasteh2024validitylearningfailuresmitigating}, and analyzing their joint impact on trajectory clustering and priority sampling effectiveness.

% \IEEEtriggeratref{16}
\bibliographystyle{IEEEtran.bst}

\typeout{}
\bibliography{01-references.bib}

\end{document}